\documentclass{article}

\PassOptionsToPackage{numbers, compress}{natbib}

\usepackage[preprint]{neurips_2024}
\usepackage{kotex}

\usepackage[utf8]{inputenc} 
\usepackage[T1]{fontenc}    
\usepackage{hyperref}       
\usepackage{url}            
\usepackage{booktabs}       
\usepackage{amsfonts}       
\usepackage{nicefrac}       
\usepackage{microtype}      
\usepackage{xcolor}         
\usepackage{graphicx}       
\usepackage{amsmath}        
\usepackage{multirow}       
\usepackage{caption}

\title{A2SF: Accumulative Attention Scoring with Forgetting Factor for Token Pruning in Transformer Decoder}

\author{%
  Hyun-rae Jo \\
  Department of Electrical and Computer Engineering\\
  Sungkyunkwan University\\
  \texttt{smp9898@skku.edu} \\
  \And
  Dongkun Shin \\
  Department of Electrical and Computer Engineering \\
  Sungkyunkwan University \\
  \texttt{dongkun@skku.edu} \\
}

\begin{document}

\maketitle

\begin{abstract}

Recently, large language models (LLM) based on transformers are facing memory bottleneck issues due to KV cache, especially in long sequence handling. Previous researches proposed KV cache compression techniques that identify insignificant tokens based on Accumulative Attention Scores and removes their items from KV cache, noting that only few tokens play an important role in attention operations. However, we have observed that the existing Accumulative Attention Score is not suitable for the transformer decoder structure. In the decoder model, the number of times the Attention Score accumulates varies depending on the order of token appearance due to the effect of masking, causing an uneven comparison between tokens. To solve this, we propose Accumulative Attention Score with Forgetting Factor (A2SF) technique, which introduces a “Forgetting Factor” in the Attention Score accumulation process. A2SF applies a penalty to the past Attention Score generated from old tokens by repeatedly multiplying the Forgetting Factor to the Attention Score over time. Therefore, older tokens receive a larger penalty, providing fairness among different ages of tokens. Through the fair comparison among tokens, we can more effectively select important tokens. We have verified the accuracy improvement through A2SF in the OPT and LLaMA models and A2SF improves the accuracy of LLaMA 2 by up to 7.8\% and 5.1\% on 1-shot and 0-shot. The code is available at~\url{https://github.com/Dirac-Notation/A2SF}

\end{abstract}

\section{Introduction}

In recent years, Transformers-based Large Language Models (LLMs)~\cite{transformer} have made significant strides. These models are now widely used in various domains, including content creation, text summarization, and chatbots. However, the computational and memory demands of these models are escalating rapidly to maintain high accuracy. Unlike the Transformer Encoder that processes the entire input simultaneously, the Transformer Decoder, which feeds the generated token back as input, needs to recalculate the Key and Value for the previously generated tokens. To address the issue posed by this auto-regressive characteristic, a Key-Value (KV) Cache is employed to store the Key and Value of the generated token in memory. The KV Cache, which retains the Key and Value for all tokens, has a memory usage that increases linearly with the sequence length. Consequently, the memory usage tend to escalate with longer sequences and larger batches. In particular, repeated memory access to load the KV Cache can create a bottleneck due to memory bandwidth limitations. In the worst-case scenario, computation may become impossible due to insufficient GPU memory. To mitigate the issue of limited GPU memory capacity, LLM optimization techniques~\cite{flexgen, pagedattention, accelerate} have proposed offloading the KV Cache to CPU memory. However, this increases data traffic between the GPU and CPU, exacerbating the bottleneck, and as the sequence lengthens, the continually expanding KV Cache can eventually exhaust CPU memory space. Therefore, a fundamental solution is required to effectively reduce the size of the KV Cache. 

\begin{figure}
  \centering
  \includegraphics[width=1.0\textwidth]{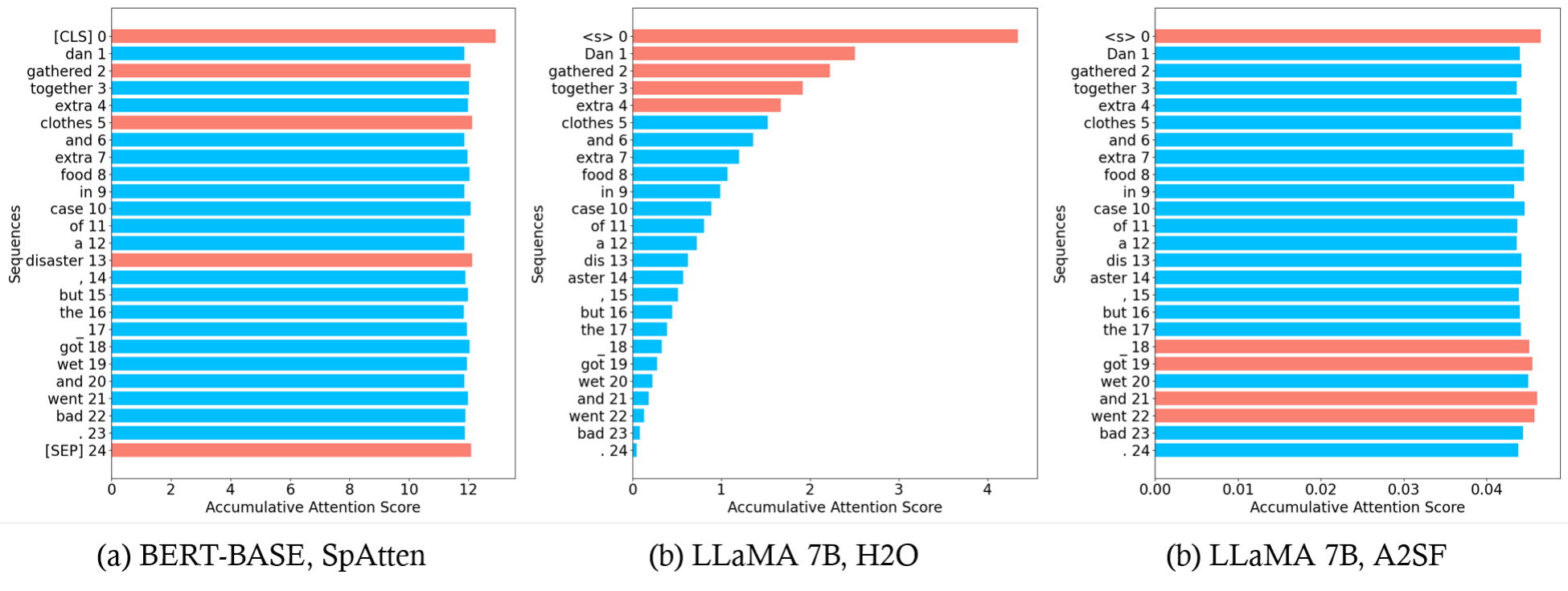}
  \caption{A2S results of encoder model and decoder model and A2SF results in a winogrande data example. SpAtten and H2O use A2S. Unlike the results in BERT, the comparison of tokens in LLaMA is not being properly conducted. The imbalance can be resolved through A2SF.}
  \label{AAS}
\end{figure}

SpAtten~\cite{spatten} illustrated that important tokens can be distinguished through the Accumulative Attention Score (A2S), which accumulates the output values of the Softmax function on a per-token basis, in Encoder models. They proposed a token pruning technique that calculates the A2S of each token while calculating a layer, removes tokens with small A2S during the execution of the next layer, and only uses the remaining tokens. This technique has the advantage of being applicable to the Pre-trained Model in a Plug-and-Play manner without incurring additional costs.

H2O~\cite{h2o} extended SpAtten's Token Pruning technique to the Transformer Decoder-based LLM model. Due to the structural differences between the Encoder model and the Decoder model, which has auto-regressive characteristics, the method of accumulating A2S across layers was modified to accumulate along the Generative step. By maintaining each A2S in each Attention-head, it was confirmed that the Token Pruning method based on A2S is also effective in LLMs. Subsequent research~\cite{notokenleft, getmorewith, keyformer} improving H2O's technique highlighted problems arising from H2O's token processing and Softmax, and proposed techniques to address them. Consequently, numerous studies are being conducted using A2S as a token selection method, which can effectively distinguish important tokens through a simple method.

Meanwhile, the Transformer Decoder comprises Masked Self-Attention, which applies a Causal Mask during the Attention operation. It masks the upper triangular area of Attention. Through this, the result of the softmax function has a probability of 0 in the masked area. This blocks the connection between the current token and the future token, indicating that the information of the future token cannot be known in the process of generating current tokens.

Despite numerous studies using A2S, they do not consider the characteristics of the above Causal Mask. The output of Masked Self-Attention always has an upper triangular area of 0, leading to an imbalance in the process of A2S depending on the order of token appearance. Tokens that were generated in the past have a lot of A2S because there are few tokens that are masked in the attention operation, and tokens that were generated recently have a small amount of accumulation because the Attention Score is 0 for all tokens before the token. Therefore, the likelihood of measuring the importance of tokens that appeared in the past is high. This problem makes it impossible to make a precise comparison about the tokens, and often selects unnecessary tokens that were generated earlier. This is a critical problem that can lower the accuracy of the model by removing important tokens.

To address this issue, we propose A2SF (Accumulative Attention Score with Forgetting Factor), which can resolve the unfairness between tokens about Accumulative Attention Score. A2SF applies a Forgetting Factor to the Attention Score to reduce the value of past Attention Scores. By increasing the number of applications of the factor over time, it gradually forgets the past and reduces its impact on the present. Therefore, since many Factors are applied to the Score of old tokens, it can resolve the imbalance with recent tokens.

We conducted experiments with various models such as LLaMA and OPT, and confirmed that A2SF shows accuracy improvement in many datasets compared to existing techniques. We were able to confirm that the accuracy of the LLaMA 2 7B model, which has a cache ratio of 0.2, increased by an average of 7.8\% in a 1-shot test and by an average of 5.1\% in a 0-shot test compared to H2O through A2SF.

\section{Related Works}

\subsection{Token pruning for Transformer Encoder}

Prior to the exploration of token pruning in Transformer Decoder models, significant research was conducted on token pruning in Transformer Encoder models such as BERT~\cite{bert}. These encoder models process all tokens simultaneously, and their primary use was to decrease computational load rather than to manage memory. 

Longformer~\cite{longformer} introduced predetermined mask for the attention operation. This included Sliding Window Attention, which only considers tokens within a certain distance from the current token, and Dilated Sliding Window Attention. These methods provided a straightforward approach to token pruning. However, since the same token pattern is applied irrespective of the input, there is a potential for accuracy loss due to the inability to consider the context.

In contrast, SpAttn proposed a token pruning technique that utilizes Accumulative Attention Score (A2S). A2S is a method for determining token importance that accumulates Attention Score, which is generated during the attention operation, in units of Key direction tokens. SpAttn further accumulates this A2S at the layer level, establishes the token importance after each layer's operation, and suggests a method to exclude tokens with low A2S from the input of the subsequent layer. This approach is based on that in human language, many tokens, such as prepositions, articles, and adverbs, carry less semantic weight.

\subsection{Token pruning for Transformer Decoder}

The popularity of generative LLM models such as GPT~\cite{gpt} based on Transformer Decoder has been on the rise recently. Consequently, token pruning research for Decoder models has emerged as a significant issue. Unlike Encoder models, Decoder models store information about previous tokens in the form of a KV Cache, thereby token pruning reduces not only the computational load but also memory usage. Therefore, the application of the token pruning technique has become an essential requirement for rapidly utilizing LLM in various environments.

Sparse Transformers~\cite{sparsetransformer} applied a fixed pattern method, similar to the previous Encoder token pruning technique, to LLM. The proposed pattern is a suitable mix of Strided Attention and Fixed Attention. However, this technique also employs a pattern that is independent of the input, so it cannot consider the context at all, leading to a decrease in accuracy.

StreamingLLM~\cite{streamingllm} identified Attention Sink, a phenomenon where the Attention Score is concentrated on the first token during the attention operation of LLM. It also demonstrated that maintaining this Sink Token plays a crucial role in preserving the performance of LLM. Utilizing this, it was confirmed that accuracy could be significantly improved merely by retaining additional Sink Tokens in the above fixed patterns.

In H2O, the intuition of unnecessary tokens was analyzed more experimentally in Decoder models. Similar to Encoder models, most of the Attention Scores are produced by a few tokens, referred to as \textit{Heavy-Hitter}. Also, when these H2 tokens were removed, the accuracy dropped significantly, proving that they play a vital role in the token processing process. And, H2O proposed to apply Accumulative Attention Score-based Token Pruning  to prune the KV Cache. However, since there are differences in the token processing process between Encoder models and Decoder models, the method of accumulating A2S across layers was changed to accumulate along the Generation Step. That is, the Attention Score calculated when each token is generated is cumulatively added to the A2S of each token, and some tokens with small A2S values are removed from the KV Cache. It also proposed head-wise Token Pruning, which maintains A2S in each head and therefore removes different Tokens, improving the existing technique of applying the same Token Pruning Mask in one layer. Through this, it was confirmed that A2S operates quite well in Decoder models.

Subsequent research uses the token selection technique of H2O, but points out that H2O merely deletes unimportant tokens. In response to this, No Token Left Behind~\cite{notokenleft} proposed to convert unimportant tokens to low bits through Quantization, and Get More with LESS~\cite{getmorewith} proposed to apply Low-rank Decomposition. Both papers propose methods about processing unimportant tokens after token selection, so they can be used together with A2SF in this paper to improve overall performance.

Keyformer~\cite{keyformer} pointed out that after Token Pruning, the denominator of Softmax changes because tokens are removed, and therefore the distribution changes. To solve this problem, it proposed to use Gumbel-Softmax~\cite{gumbel}, which can flatten the distribution. This technique is also compatible with A2SF because they use A2S after Gumbel-Softmax.

Other research~\cite{dcp, dmc} introduces additional learning in the process of selecting important tokens. However, LLM requires a lot of computation and memory during the learning process due to its size, which can be challenging to apply in environments with insufficient computing power and memory.
 
\section{Method}

\subsection{Accumulative Attention Score}
The A2S used in the Encoder model is defined as follows:

\begin{equation}
A^l_{k} = \sum_{i=1}^{l} \sum_{h=1}^{H} \sum_{q=1}^{N} S^{i,h}_{q,k}
\end{equation}

Here, $A^l_{k}$ is the A2S for the $k$th token in the $l$th layer, $N$ is the length of the entire sequence, $H$ is the total number of heads, and $S^{i,h}_{q,k}$ is the Attention Score calculated for the $q$th query and the $k$th key in the $h$th head of the $i$th layer. This A2S is calculated from the front layers as the operation progresses, allowing each token's importance to be set in that layer. Subsequently, tokens with low importance can be removed from the input of the next layer, reducing the amount of computation.

In H2O, the following changes were made to apply A2S to the Decoder model:

\begin{equation}
A^{l,h}_{n,k} = \sum_{q=1}^{n} S^{l,h}_{q,k}
\end{equation}

Here, $k<n$, and by changing the overall accumulation direction to the Generation step instead of the layer, separate scores were maintained at the layer and head level. The above $l$ and $h$ mean the $h$th head of the $l$th layer. Then, $A^{l,h}_{n,k}$ is the A2S of the $k$th token in the Generation Step where the $n$th token is generated, and $S^{l,h}_{q,k}$ is the Attention Score of the $k$th token in the Step where the $q$th token is generated. Subsequently, the KV Cache of tokens with small A2S values is removed in the next Generation Step.

However, due to the introduction of the Casual Mask of Masked Self-Attention, the $k$th token does not perform Attention operations with the previous tokens:

\begin{equation}
S^{l,h}_{q,k} = 0, \forall q < k
\end{equation}

Therefore, the actual A2S applied to the Decoder model is as follows:

\begin{equation}
A^{l,h}_{n,k} = \sum_{q=k}^{n} S^{l,h}_{q,k}
\end{equation}

Through this, $n-k$ Attention Scores are accumulated in the A2S of the $k$th token, and $n-k-10$ Attention Scores are accumulated in the A2S of the $k+10$th token. That is, there is a difference of $10$ Attention Scores between these two tokens. The $Softmax$ operation does not output negative values due to its characteristics, so if more accumulations occur, the value is likely to be large, which means that the A2S of the token that appeared first is high and is less likely to be removed during Pruning.

\subsection{Motivation}

Figures~\ref{AAS}(a) and (b) display the results of applying A2S to BERT and LLaMA, respectively. The top 5 tokens in terms of importance are highlighted in red. In the BERT model (Figure~\ref{AAS}(a)), since the Casual Mask was not applied, the Attention Score is evenly distributed among all tokens. This is because the number of accumulations is the same for all tokens, allowing them to be compared fairly based solely on the difference in Attention Score. Conversely, in the case of LLaMA 7B (Figure~\ref{AAS}(b)), the importance of the initial tokens is abnormally high due to the effect of the Casual Mask. Also, the importance aligns with the order of generation, which can be considered absolute due to the number of accumulations according to the order of generation rather than the difference in Attention Score. Furthermore, even if the sequence lengthens, there is no change in the previously accumulated Score, making the progress of the sentence and the selected token irrelevant, and the token that was initially generated is always selected.

This differs from human language processing. Human language is written in one paragraph or one sentence, and after moving on, only the important keywords from the previous paragraph are remembered. This implies that humans naturally forget unnecessary parts that appeared in the past during the process of language processing. We derived from this point that the past Attention Score should also be forgotten over time. Also, we adopted a method of setting a Forgetting Factor and multiplying it exponentially based on the well-known human Forgetting study, Ebbinghaus's Forgetting Curve~\cite{forgetcurve}. This reflects the fact that the simplified form of the Curve is exponential.

\subsection{Accumulative Attention Score with Forgetting Factor}

\begin{figure}
  \centering
  \includegraphics[width=0.8\textwidth]{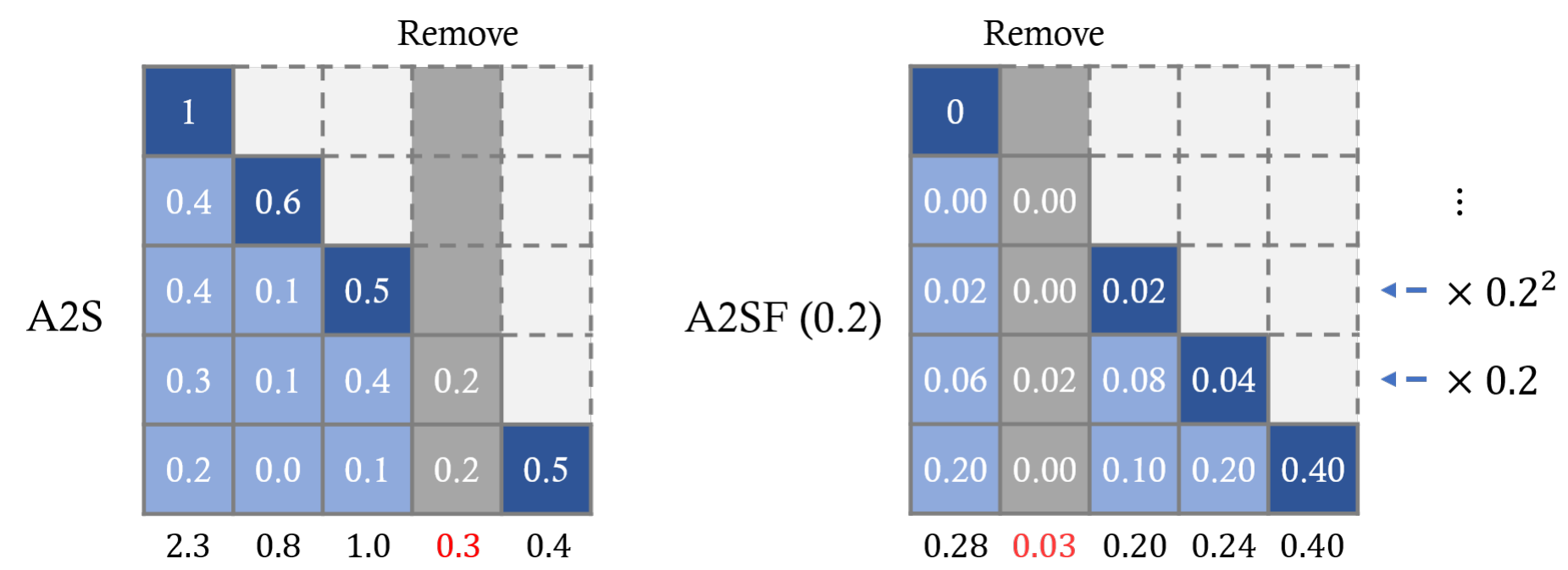}
  \caption{Toy examples illustrate the Accumulative Attention Score and the Accumulative Attention Score with Forgetting Factor.}
  \label{Method}
\end{figure}

We propose to modify the Accumulative Attention Score by introducing a Forgetting Factor $\alpha$ as follows:

\begin{equation}
A^h_{n,k} = \sum_{q=1}^{n} \alpha^{n-q} \times S^h_{q,k}
\end{equation}
\begin{equation}
A^h_{n,k} = S^h_{n,k} + \alpha \cdot S^h_{n-1,k} + \alpha^2 \cdot S^h_{n-2,k} + \ldots + \alpha^{N-k} \cdot S^h_{k,k}
\end{equation}

Here, $\alpha$ is a float number satisfying $0 < \alpha < 1$. Because the Forgetting Factor is repeatedly multiplied by the Attention Score calculated in the past Generation Step, its value converges to 0. Therefore, even if many values are accumulated, the value becomes smaller, resolving the imbalance with tokens with fewer accumulations. This can be seen in Figure~\ref{Method}. In the case of the second token, it outputs a low Attention Score after its first self-attention and becomes an unnecessary token, but in A2S, it can be seen that it is not deleted due to the influence of the past. On the other hand, in A2SF, the scores in the upper left, which are the Attention Scores created in the past from the old tokens, can be seen to gradually converge to 0 due to the influence of the Forgetting Factor. Through this, it can be confirmed that the token that is most unnecessary at the current point is removed, not the token that was removed with fewer accumulations. The results in actual data can also be confirmed through Figure~\ref{AAS}(c). Unlike Figure~\ref{AAS}(b) where A2S was applied, it can be seen that the distribution of scores between tokens is even. Through this, tokens can be compared more fairly in terms of importance, and more accurate results can be output by selecting actually important tokens.

Also, A2SF has the potential for tuning considered for the dataset by setting the value of $\alpha$. If a value close to 1 is assigned to $\alpha$, convergence occurs slowly, so the Attention Score that occurred in the past can still affect the present, but if $\alpha$ is close to 0, it converges quickly, so the Attention Score calculated recently is used to determine the importance of the token, which means comparing tokens using only recent trends. In other words, depending on the characteristics of the dataset, the degree of influence of the past on the present can be adjusted, and it means that the optimal accuracy for the situation can be achieved by setting an appropriate $\alpha$.

\section{Experiments}

\subsection{Experimental setup}

For the accuracy measurement experiment of A2SF, we used the LLaMA-2-7B~\cite{llama2}, LLaMA-7B~\cite{llama}, OPT-6.7B, OPT-2.7B~\cite{opt} models, and evaluated the Commonsense-reasoning performance with the dataset of OpenbookQA~\cite{openbookqa}, Winogrande~\cite{winogrande}, PiQA~\cite{piqa}, COPA~\cite{copa}, MathQA~\cite{mathqa}, ARC-easy, ARC-challenge~\cite{arc}. We used lm-eval-harness~\cite{lmeval}(v0.4.0) for evaluation. The experiment was conducted with an FP16 model in an RTX 3090 environment. 
In H2O, Local Attention and token selection through A2S are used together, and each ratio is set to half of the cache ratio. Since A2SF includes considering recent Attention Score trends in the algorithm, we did not allocate a separate local cache ratio and allocated all to selective cache ratio.

\subsection{Model accuracy}

\begin{figure}
  \centering
  \includegraphics[width=1.0\textwidth]{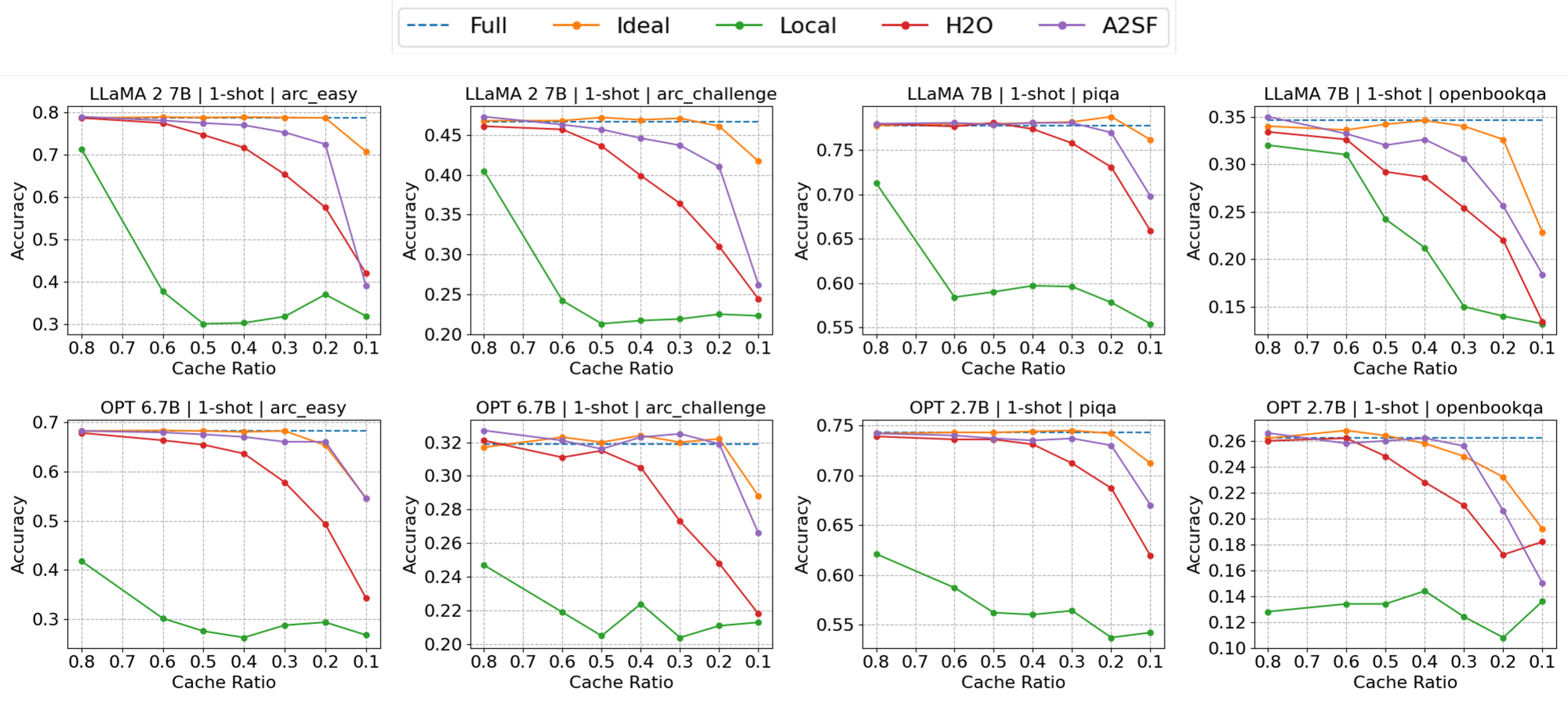}
  \caption{Comparison of accuracy between Full cache, Local Attention, and H2O}
  \label{Plot}
\end{figure}

We measured the accuracy of a total of 4 models in 1-shot and 0-shot tasks for cache ratios in the range of $[0.1, 0.8]$ across 7 datasets. Among them, the results for the 1-shot task are summarized in Figure~\ref{Plot}. Through this, we were able to draw the following conclusions: (1) In most cases, A2SF outperforms H2O in terms of accuracy. (2) At a cache ratio of 0.4 or higher, it can approach the accuracy of Ideal under many conditions. (3) There are cases where the limit of accuracy that can be achieved with a small number of tokens is low, depending on the dataset. The full experimental results of LLaMA 2 7B are in Table~\ref{Accuracy}. We were able to confirm an accuracy improvement of 7.8\% in the 1-shot environment and 5.1\% in the 0-shot environment compared to H2O.

\begin{table}
    \centering
    \setlength{\tabcolsep}{6pt}
    \renewcommand{\arraystretch}{1.05}
    \begin{tabular}{cccccccccc}
    \toprule[1.5pt]
    \multirow{2}{*}{Shots}  & \multirow{2}{*}{Method}     & \multicolumn{7}{c}{Dataset}       & \multirow{2}{*}{Average} \\
    \cline{3-9}
                            &       & OBQA	& WG	& PiQA	& COPA  & MathQA  & ARC-E & ARC-D                \\
    \hline \hline
    \multirow{4}{*}{1-shot} & Full  & 0.362 & 0.691 & 0.774 & 0.850 & 0.303   & 0.786 & 0.467 & 0.605        \\
                            & Local & 0.122 & 0.487 & 0.574 & 0.630 & 0.207   & 0.370 & 0.225 & 0.374        \\
                            & H2O   & 0.224 & 0.511 & 0.726 & 0.660 & 0.202   & 0.575 & 0.310 & 0.458        \\
                            & A2SF  & 0.268 & 0.526 & 0.756 & 0.820 & 0.251   & 0.724 & 0.410 & 0.536        \\
    \hline
    \multirow{4}{*}{0-shot} & Full  & 0.324 & 0.688 & 0.781 & 0.860 & 0.282   & 0.763 & 0.423 & 0.589        \\
                            & Local & 0.130 & 0.487 & 0.540 & 0.640 & 0.194   & 0.330 & 0.215 & 0.362        \\
                            & H2O   & 0.152 & 0.512 & 0.676 & 0.570 & 0.210   & 0.419 & 0.253 & 0.399        \\
                            & A2SF  & 0.176 & 0.526 & 0.711 & 0.620 & 0.221   & 0.586 & 0.311 & 0.450        \\
    \bottomrule[1.5pt]
    \end{tabular}
    \caption{Comparison of the performance of existing techniques and A2SF for different number of shots in LLaMA 2 7B, with a Cache Ratio of $0.2$}
    \label{Accuracy}
\end{table}

\subsection{Token selection}

\begin{figure}[h]
  \centering
  \includegraphics[width=0.9\textwidth]{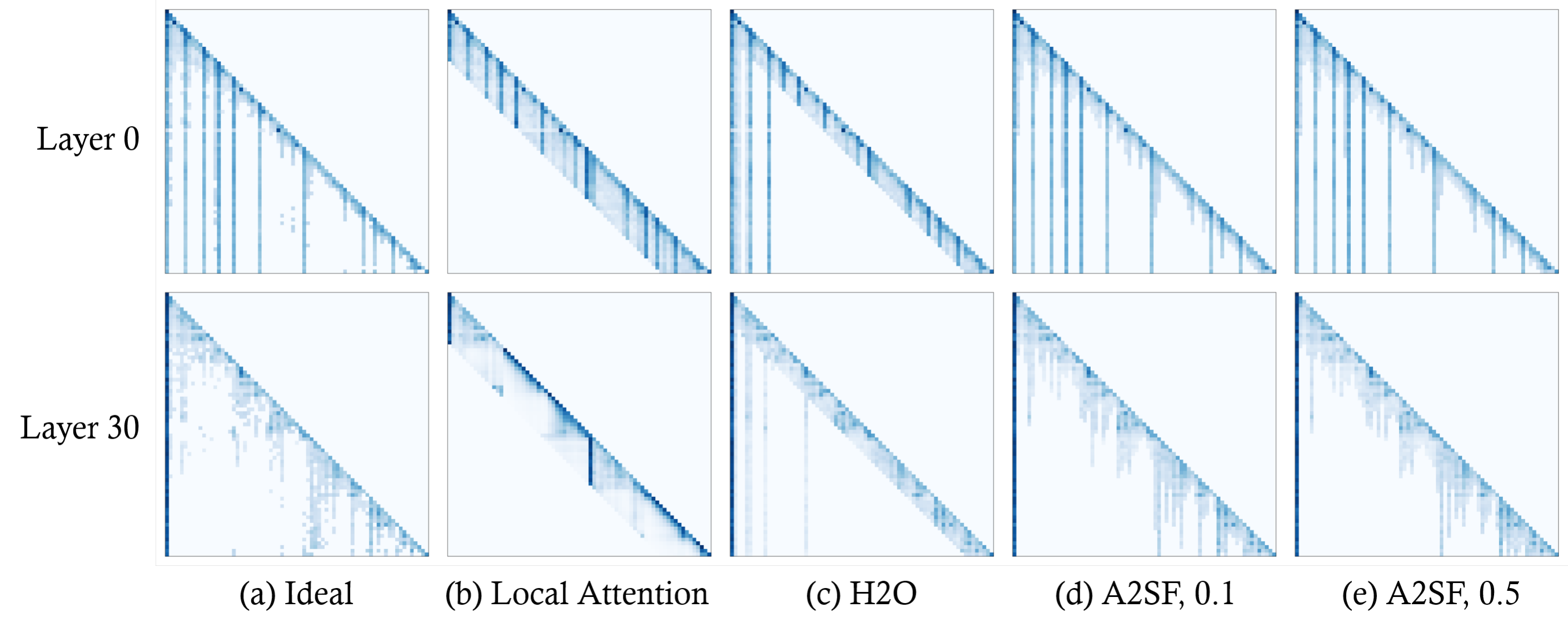}
  \caption{Attention Score after applying each technique. The intensity of the color is directly proportional to the value it represents; a darker color signifies a higher value.}
  \label{Mask}
\end{figure}

Figure~\ref{Mask} depicts the Attention Score after applying each technique with a cache ratio of 0.2. The \textit{Ideal} in Figure~\ref{Mask}(a) calculates the Attention Score of all tokens at each sequence generation step (each row), selects tokens with large values, and considers all tokens in the subsequent step without removing tokens with small values. This represents the Mask with the largest Attention Score given the Cache Size. However, since tokens are not removed, there is no compression effect, making it an ideal method. Therefore, the closer the result of applying the Token Pruning technique is to this Ideal Mask, the closer it is to achieving ideal accuracy.

Local Attention (Figure~\ref{Mask}(b)) exhibits a lower mask similarity because it only considers tokens of a certain size based on the most recent tokens. Also, due to the absence of an Attention Sink, a significant change in the distribution of Softmax can be observed. H2O (Figure~\ref{Mask}(c)) creates a $\lambda$-shaped Mask due to the issue of high scores of initial tokens, which differs from the Ideal Mask. In contrast, when A2SF was applied, a Mask is similar to the Ideal Mask compared to existing techniques. Moreover, the Attention Sink Token, which can significantly impact accuracy, is selected without any issues. This is because even when the Forgetting Factor is applied, the Sink Token is continuously calculated with a large value, allowing it to maintain a high value compared to other tokens.

In Table~\ref{Similar}, the average cosine similarity between the Attention Score of techniques and the Ideal is presented. Local Attention, where the Attention Score distribution is disrupted, shows a very low similarity. Although the Sink Token is included, H2O exhibits a relatively low similarity because the area in the middle of the $\lambda$ disappears. On the other hand, the Mask created through A2SF has an average similarity close to 1. This result explains why the model's accuracy increases when A2SF is applied.

\begin{table}
    \centering
    \setlength{\tabcolsep}{10pt}
    \renewcommand{\arraystretch}{1.05}
    \begin{tabular}{cccccc}
    \toprule[1.5pt]
    \multirow{2}{*}{Method} & \multicolumn{4}{c}{Dataset}    & \multirow{2}{*}{Average} \\
    \cline{2-5}
                            & WG    & PiQA  & OBQA  & ARC-E  &                      \\
    \hline \hline
    Local                   & 0.320 & 0.315 & 0.317 & 0.318  & 0.318                \\
    H2O                     & 0.960 & 0.970 & 0.968 & 0.970  & 0.967                \\
    A2SF 0.1                & 0.990 & 0.992 & 0.990 & 0.991  & 0.991                \\
    A2SF 0.5                & 0.988 & 0.991 & 0.987 & 0.988  & 0.989                \\
    \bottomrule[1.5pt]
    \end{tabular}
    \caption{Average cosine similarity between the Attention Score generated by each method and the Attention Score applied with Ideal Pruning. Average cosine similarity between the Attention Score generated by each method and the Attention Score applied with Ideal Pruning.}
    \label{Similar}
\end{table}

\subsection{Discussion}

\subsubsection{Determining the optimal Forgetting Factor value}

\begin{figure}
\centering
\includegraphics[width=1.0\textwidth]{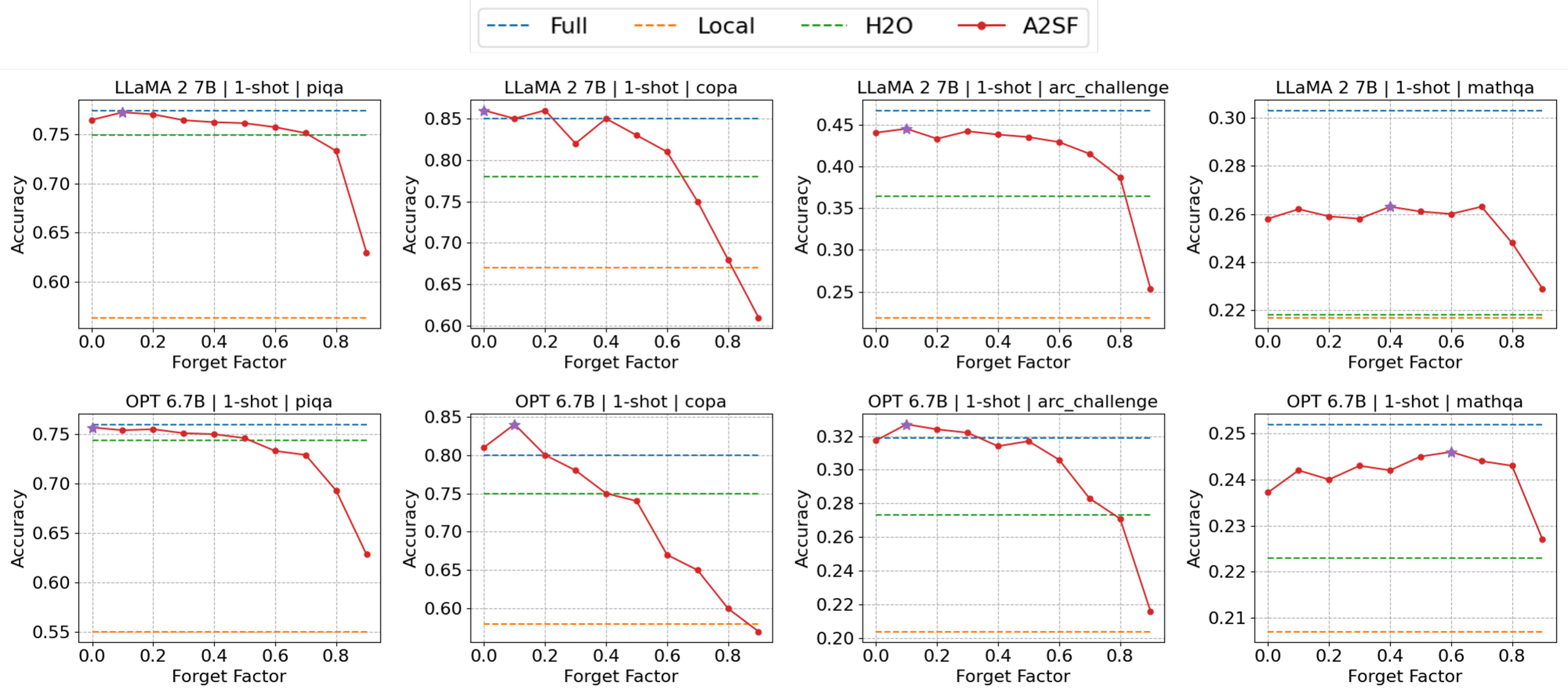}
\caption{Factor graph}
\label{Factor}
\end{figure}
Figure~\ref{Factor} illustrates the accuracy achieved when applying a Forgetting Factor within the range of $[0.0,0.9]$, with a fixed cache ratio of $0.3$. Generally, the peak accuracy is observed within the Forgetting Factor range of $[0.1,0.3]$, and the accuracy tends to decline as the Factor increases. A relatively low Forgetting Factor implies that the model considers a shorter history. This observation suggests that when applying the A2S to the Transformer Decoder model, an extensive historical context can have a detrimental effect.

However, in the MathQA dataset, there were instances where a relatively high Forgetting Factor resulted in high accuracy, and in some cases, even the highest accuracy. This discrepancy can be attributed to the differences between the datasets. Here are some examples of sequences included in each dataset:
\begin{itemize}
    \item OpenbookQA
        \begin{itemize}
            \item If you wanted to make a necklace, how long would you have to wait for the materials to appear inside the Earth? Millions of years
        \end{itemize}
    \item PiQA
        \begin{itemize}
            \item Question: What ingredient is left out of fluffy slime that is normally in regular slime? Answer: Borax
        \end{itemize}
    \item MathQA
        \begin{itemize}
            \item Question: In a 160 meters race, a beats b by 56 m or 7 seconds. What is a’s time over the course? Answer: 22 seconds
        \end{itemize}
\end{itemize}
Compared to other datasets, a distinctive feature of MathQA is its clear problem structure due to the use of numbers. In other words, the distinction between necessary and unnecessary tokens is evident, suggesting that it could be beneficial to consider a longer history and prevent the removal of important tokens.

\subsubsection{Influence of history}

\begin{figure}
  \centering
  \includegraphics[width=1.0\textwidth]{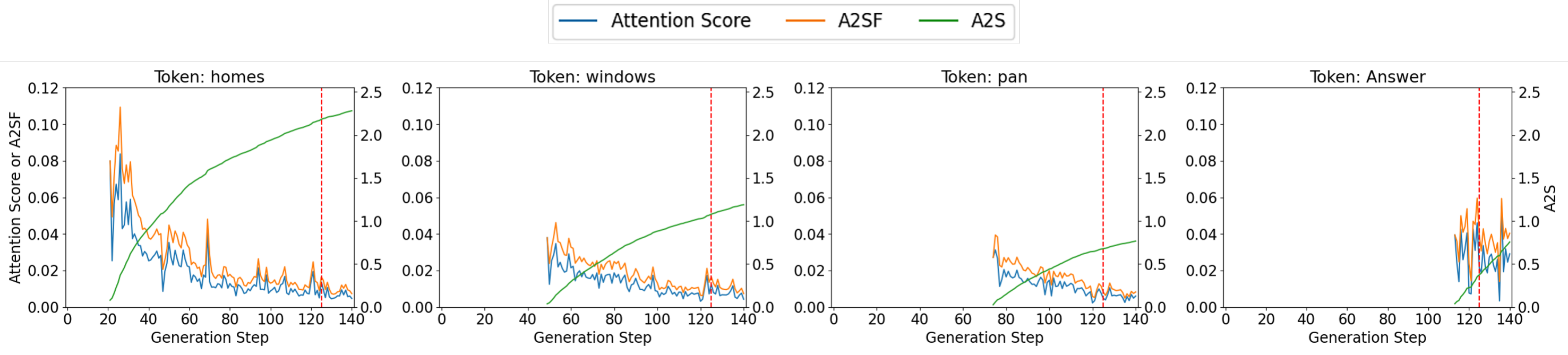}
  \caption{Attention Score, A2S and A2SF of the important tokens according to the Generation Step. The starting point of the graph differs because each token is generated at different steps.}
  \label{Token}
\end{figure}

In Figure~\ref{Factor}, we observed instances where the highest accuracy was attained when the Forgetting Factor was set to $0.0$. A Forgetting Factor of $0.0$ implies that the model does not consider history at all, and it determines the importance of a token solely based on the Attention Score from the immediate preceding sequence. This approach demonstrated higher accuracy across all cases compared to H2O, which takes into account all history. This suggests that for the Decoder model, it is more beneficial to disregard history entirely rather than considering an excessive amount of it. However, we generally observed higher accuracy when the model considered recent history through a low Forgetting Factor in the range of $[0.1, 0.3]$, indicating that history can serve as a supplementary factor.

Figure~\ref{Token} illustrates the Attention Score generated by the primary tokens as the Generation Step progresses, along with the A2S and A2SF calculated from it. This figure reveals that the remaining tokens, excluding the Attention Sink, do not consistently have a large Score, even if they are major tokens. In this process, history plays a role in preventing a token, which had a large Score, from being eliminated during a certain Step by compensating the value when it outputs a small Score, or awarding extra points when it continuously outputs a large Score. Since A2S maintains the accumulated value from the past, a token that was initially created carries a high level of importance, even if the Score generated over time is small. If Token Pruning is performed at the point indicated by the red line in Figure~\ref{Token}, \textit{Answer} is removed despite its high Attention Score due to its small accumulated value. In contrast, A2SF accounts for the reduced Score of past tokens, so the importance of the \textit{Answer} remains high.

\section{Conclusion}

We propose a novel method, A2SF (Accumulative Attention Score with Forgetting Factor), to rectify imbalances in the Accumulative Attention Score utilized for token importance evaluation in Transformer Decoder-based models. A2SF addresses the skewed importance scoring induced by the causal masking in Masked Self-Attention, which disproportionately amplifies the significance of early sequence tokens. A2SF incorporates a Forgetting Factor to mitigate this historical bias in Attention Scores, thereby ensuring a more equitable distribution of token importance across the sequence length. Comprehensive experiments on a variety of models and datasets demonstrate that A2SF surpasses existing methods, enhancing accuracy without necessitating additional model retraining. Furthermore, given that A2SF is a token selection technique, it holds potential to enhance the performance of numerous ongoing KV Cache processing algorithms when integrated with A2SF.

\section{Limitation}

We applied a uniform value for the Forget Factor across all heads, layers, and tokens. However, the characteristics of these elements can vary significantly. For instance, tokens can represent diverse features such as nouns, verbs, and so on. Therefore, applying the same rate of forgetting to all elements may not yield optimal results. For example, when summarizing content, it could be beneficial to retain as much information as possible about unique nouns that form the core of a paragraph. To address this, we could devise methods to determine an appropriate Forget Factor based on the form of the Attention Score generated during previous generation steps. Alternatively, we could set a distinct Forget Factor for each layer, head, and token by understanding their characteristics in advance.

\bibliographystyle{unsrt}
\bibliography{ref}

\end{document}